\documentclass{article}
\usepackage{PRIMEarxiv}
\usepackage{subcaption}
\usepackage{placeins} 
\usepackage{float}
\usepackage{appendix}
\usepackage{multirow}
\usepackage{natbib}
\usepackage[utf8]{inputenc} 
\usepackage[T1]{fontenc}    
\usepackage{hyperref}       
\usepackage{url}            
\usepackage{booktabs}       
\usepackage{amsfonts}       
\usepackage{nicefrac}       
\usepackage{microtype}      
\usepackage{lipsum}
\usepackage{fancyhdr}       
\usepackage{graphicx}       
\graphicspath{{media/}}     
\usepackage{float}
\usepackage{tcolorbox}
\newtcolorbox{boxtext}{
  colback=gray!10,
  colframe=gray!80,
  boxrule=0.8pt,
  arc=4pt,
  left=6pt,
  right=6pt,
  top=6pt,
  bottom=6pt,
}

\usepackage{amsmath}
\title{Large Language Models as `Hidden Persuaders': \\ Fake Product Reviews are Indistinguishable \\ to Humans and Machines\thanks{\textit{This manuscript is a preprint and currently under review.}}}

\author{
  Weiyao Meng\textsuperscript{1,3},
  John Harvey\textsuperscript{1},
  James Goulding\textsuperscript{1},
  Chris James Carter\textsuperscript{2},
  Evgeniya Lukinova\textsuperscript{1},
  Andrew Smith\textsuperscript{1},\\
  Paul Frobisher\textsuperscript{3},
  Mina Forrest\textsuperscript{3},
  Georgiana Nica-Avram\textsuperscript{1} \\
  \\
  \textsuperscript{1}N/LAB, Nottingham University Business School, University of Nottingham, UK\\
  \textsuperscript{2}Haydn Green Institute for Entrepreneurship and Innovation, University of Nottingham, UK\\
  \textsuperscript{3}Strategic Innovation Ltd, UK
}
\begin{document}
\maketitle










\begin{abstract}
Reading and evaluating product reviews is central to how most people decide what to buy and consume online. However, the recent emergence of Large Language Models and Generative Artificial Intelligence now means writing fraudulent or fake reviews is potentially easier than ever. Through three studies we demonstrate that (1) humans are no longer able to distinguish between real and fake product reviews generated by machines, averaging only 50.8\% accuracy overall – essentially the same that would be expected by chance alone; (2) that LLMs are likewise unable to distinguish between fake and real reviews and perform equivalently bad or even worse than humans; and (3) that humans and LLMs pursue different strategies for evaluating authenticity which lead to equivalently bad accuracy, but different precision, recall and F1 scores – indicating they perform worse at different aspects of judgment. The results reveal that review systems everywhere are now susceptible to mechanised fraud if they do not depend on trustworthy purchase verification to guarantee the authenticity of reviewers. Furthermore, the results provide insight into the consumer psychology of how humans judge authenticity, demonstrating there is an inherent `scepticism bias' towards positive reviews and a special vulnerability to misjudge the authenticity of fake negative reviews. Additionally, results provide a first insight into the ‘machine psychology’ of judging fake reviews, revealing that the strategies LLMs take to evaluate authenticity radically differ from humans, in ways that are equally wrong in terms of accuracy, but different in their misjudgments.
\end{abstract}



\section{Introduction}
Each year, an estimated 2.5 billion people buy something online. You are likely one of those people. Most e-commerce stores now routinely display product reviews. These are typically short, instructive paragraphs of text that detail the observations and judgments of other consumers, pointing to the relative benefits and limitations of the respective item and the associated consumption experience. There is clear evidence that product reviews are persuasive and influence purchase decisions \citep{duan2008online, zhu2010impact, forman2008examining, chevalier2006effect}. Yet, the relative scale of fraud across online product reviews remains unknown - as is how many of the world's 2.5 billion online consumers base their decisions on deliberately manipulated or fraudulent reviews.

The urgency of the issue continues to grow due to the widespread adoption of new \textit{Generative Artificial Intelligence} techniques (or `GenAI') for creating text, underpinned by recent advances in Large Language Models (LLMs). In many domains, resulting tools now have the capability to produce online posts that have the appearance of being entirely human, despite entirely artificial construction. The credibility of GenAI writing is so striking that it threatens to upturn entire industries, with the most widely described threat of artificial intelligence (AI) - beyond superintelligence destroying humanity \citep[e.g.,][]{bostrom2017superintelligence, tegmark2018life} - focusing on technology supplanting the workforce of human labour.  Whether replacing lawyers, medical doctors, or academic professors, there is a growing concern that technology will render a broad class of knowledge workers redundant \citep{brodeur2024superhumanperformancelargelanguage, schneiders2024objection, sikander2023chatgpt}. 

What has received comparatively less scrutiny is how these technologies will inevitably be used for persuasion. Far from eliminating the role of the marketer, GenAI will more likely enhance the arsenal of persuasive tools available to manipulate consumers. LLMs provide an instantaneous, multilingual, and highly accessible means to persuade consumers through the creation of fake product reviews. It is this precise issue - of generative artificial tools being used by unscrupulous marketers to generate hyper-persuasive but false narratives - which we aim to examine. In short, these tools potentially make the act of industrial fraudulence via product reviews a trivial task.

As part of the broader wave of GenAI innovations, LLMs are increasingly embedded into consumer-facing technologies - a process that has been described as a consumer-centric disruptive force that is reshaping the modern economy and accelerating the shift toward automation and data-driven decision-making \citep{beheshti2024overview}. In contexts such as customer service, branding, and digital marketing, LLMs now play an active role in content creation and recommendation \citep{lee2022parasocial, vernuccio2023delving, li2023friendship, ford2023ai, osadchaya2024chatgpt, cui2024build, ferraro2024paradoxes}; and despite often not obvious to end users, an increasingly large fraction of Web-based content is written by machines \citep{thompson2024shocking}. This is happening now, not in some hypothetical future. However, given the surreptitious nature of content designed to persuade, it is difficult to reliably estimate its prevalence. The idea of marketers as '\textit{hidden persuaders}' capable of leveraging academic insight into psychological and social sciences to create surreptitious manipulation practices has a long history \citep[at least as far back as][]{packard1957hidden}. We will not repeat such critique of marketing here, other than to suggest that the emergence of LLMs has created the potential for marketers to generate and deploy an army of false consumer advocates in their attempts as hidden persuaders. In this article, we present a series of studies to show that this shift has potential to be even more insidious. 
Understanding the manipulative intent of adverts requires a certain level of literacy, 
and this knowledge is often not fully realised in most people until late adolescence \citep{john1999consumer, boush1994adolescent}. But in the case of identifying false consumer narratives, is literacy enough to defend ourselves?

\subsection{Background and Research Aims}



Emerging evidence suggests that consumers often struggle to distinguish AI-generated from human-written content across a range of domains, including news, product descriptions, and service reviews \citep{clark2021all, salminen2022creating, kovacs2024turing, hatch2025eliza}. Where LLMs have previously been shown to pass the ‘Turing Test’ (i.e. a test of a machine's ability to exhibit intelligent behaviour equivalent to that of a human), this is less obvious for the product review format. Product reviews, however, are relatively unusual compared to many other forms of narrative where LLMs have demonstrated effectiveness in generating. Unlike computer programming, disease diagnosis, the legalese of lawyers, or news articles, product reviews are not so dependent on highly technical language or syntax, which can easily fool a reader. Instead, they often contain half-formed ideas, are replete with typographic errors, include non-sequiturs, and are dependent on the local context in which a product was purchased, delivered, consumed or disposed of. If, as it is often said, that ‘to err is human’, the same cannot be said of LLMs, which are designed to generate flawless text. The imperfections of human-generated product reviews are unusual compared to both typical human-edited or AI-generated text. Therefore, we are interested in asking whether it is still possible for either people or indeed LLMs to distinguish false AI generated reviews. This aim frames the three studies that follow.

While prior work has explored the detection of AI-generated content across various domains, there remains limited empirical evidence directly comparing the performance and the underlying reasoning of humans and state-of-the-art LLMs in judging product reviews. To address this gap, we aim to advance understanding of how both humans and LLMs assess the authenticity of product reviews, particularly when the content is generated by LLMs, via three studies, each correspond to the following respective research questions:

\begin{itemize}
\item \textit{Research Question 1: Can humans distinguish between real and fake reviews generated by large language models?}
Through an experimental online study, we evaluate human participants and their ability to distinguish between real and LLM-generated reviews. We additionally investigate how accuracy and self-reported confidence at identifying fakes varies across individuals according to demography and experience. \\

\item \textit{Research Question 2: Can Large Language Models distinguish between real and fake reviews?}
We extend the experimental judgment task to evaluate the detection capabilities of state-of-the-art LLMs, benchmarking their performance against human accuracy.\\  

\item \textit{Research Question 3: What makes a fake review hard to distinguish?}
The final study provides a comparative analysis of the underlying heuristics used by both humans and LLMs when making authenticity judgments.  Using a range of measures to evaluate binary classification performance, we demonstrate the respective strategies and heuristics used to inform judgment for both humans and machines. 
\end{itemize}

This paper makes two main contributions to the literature on understanding consumer behaviour and detecting AI-generated content.
Firstly, the study provides the first systematic empirical benchmark comparing human participants and state-of-the-art LLMs in their ability to distinguish between real and AI-generated product reviews. The findings reveal that both groups perform only marginally above chance, underscoring fundamental challenges in current detection capabilities. This contributes empirical clarity to growing concerns that GenAI can produce content that is not only persuasive but also effectively indistinguishable from authentic, human-generated text. More importantly, this highlights an imbalance in the generative–detective capacities of LLMs, in that they can easily produce synthetic content, but struggle to accurately detect it as such. This imbalance represents a key dimension of GenAI’s ‘dark side’, especially in consumer-facing environments where trust is critical.

Secondly, this paper identifies the cognitive and computational heuristics underlying authenticity judgments.
By analysing which textual features are associated with detection performance and alignment, this paper uncovers the divergent heuristics used by humans and LLMs. Specifically, the study reveals that humans tend to rely on intuitive cues and are biased by ‘too-good-to-be-true’ judgments, which we describe as a form of ‘scepticism bias’ towards positive reviews. These cues are often flawed and easily manipulated, supporting concerns raised in prior theoretical work about the cognitive limitations of human detection. In contrast to humans, we show that LLMs exhibit a different form of strategy when judging reviews, which reveals itself as being biased toward believing that most reviews are real reviews. We define this trait as a ‘veracity bias’, i.e., a tendency to falsely believe in the veracity of reviews.
LLM judgments are shown to rely on superficial textual features, such as review length. These findings highlight key vulnerabilities in both humans and LLMs and offer insight into how future ‘fake review’ generations could exploit these blind spots. 

The remainder of the paper is structured as follows: section \ref{sec:sec2} reviews related works on human judgment of review authenticity, AI-based fake content detection, and their differences in evaluation strategies; sections \ref{sec:sec3}, \ref{sec:sec4}, and \ref{sec:sec5} present three empirical studies, followed by a general discussion of the findings and their implications in Section \ref{sec:sec6}. Finally, section \ref{sec:sec7} concludes the paper by outlining its limitations and identifying directions for future research.

\section{Related works}\label{sec:sec2}
Online reviews are not only consulted in digital shopping environments but also in-store during showrooming, as a form of extended cognition \citep{smith2025consumer}. This reflects the logic of electronic word-of-mouth (eWOM), whereby consumers seek peer feedback to reduce perceived risk and uncertainty in purchase decisions \citep{hu2008online}.  The increasing reliance on such reviews and their persuasive power in consumer choice underscores the importance of ensuring their authenticity, particularly as AI-generated reviews become more prevalent. In 2024, the UK government announced a ‘ban’ on fake online reviews as part of a broader initiative to protect consumers, estimating that misleading endorsements and hidden fees cost consumers over £2.2 billion annually \citep{news2025}. This highlights the urgency of addressing review authenticity not only in academic research but also in policy action.

Outside of AI-generated fake reviews in e-commerce environments, there is a broad literature on authenticity and deception detection, spanning various content formats - research that identifies a range of cognitive and linguistic heuristics used in contexts such as news, social media, and human-written reviews, which inform how consumers perceive authenticity in product reviews. As such this literature review draws from a broad base of related works, presenting existing theories and highlighting potential cues relevant to fake review detection, structuring around three key areas: (1) how consumers judge authenticity and what the psychological and textual cues they rely on when judging reviews; (2) how AI systems, particularly LLMs, process and evaluate review authenticity; and (3) how human and machine judgment strategies differ in this context.

\subsection{Consumer judgment of review authenticity}

\subsubsection{Cognitive models}
According to the Heuristic-Systematic Model \citep{chaiken1980heuristic}, humans rely on two modes of information processing. One is a systematic route, which involves systematic analysis of message content, entailing a deliberate attempt to assess credibility through a broad range of cues, such as message consistency, author identity, or platform context. The other is the heuristic route, which depends on mental shortcuts and surface cues, such as perceived authority of the source, the length of the message, and the degree of social consensus.

Both analytic and heuristic strategies can operate simultaneously in credibility evaluation \citep{metzger2015psychological, metzger2007making}, but are heavily influenced by cognitive heuristics. This is reflected in studies finding that people are often biased to believe in the validity of information, and ‘go with their gut’ and intuitions instead of deliberating \citep{ecker2022psychological}. This might be because heuristic processing is usually faster, relying on surface-level characteristics (e.g., familiarity and tone) or intuitive judgments. According to the Limited Capacity Model \citep{lang2000limited} and the Prominence-Interpretation Theory \citep{fogg2003users}, individuals selectively attend to only certain salient features when evaluating information due to limited cognitive resources. 

The self-confirmation (personal opinion confirmation) heuristic is particularly relevant \citep{metzger2015psychological}, whereby people are more likely to believe information that aligns with their prior beliefs and dismiss information that contradicts those beliefs - regardless of how well-reasoned, well-sourced, or comprehensive it may be. Beyond credibility, researchers have identified three dimensions of authenticity perception: historical, categorical, and value-based \citep{newman2019psychology}. These cues are particularly relevant to the self-confirmation heuristic. That is, individuals tend to judge authenticity based on whether a piece of content fits their pre-existing beliefs about what realness should be. These judgments are inherently subjective and may be shaped by prior knowledge and expectations. 

\subsubsection{Text-based heuristics}
Human judgment of fake content has been studied across various domains and the heuristics people rely on often differ depending on the content type. In the context of fake news detection, for instance, \citet{damstra2021does} found that individuals draw on a range of cues, including ideological bias, emotional tone, verifiability, and headline structure, as well as linguistic features such as lexical diversity, capitalisation, pronoun usage, informal language, and punctuation patterns.

In the context of online consumer reviews, people tend to rely more heavily on textual and stylistic features. In \citet{chevalier2006effect}, reviews perceived as detailed or specific are often seen as more credible. Additionally, the presence of both positive and negative comments within a single review or review set tends to enhance perceived authenticity, as balanced feedback appears more genuine than uniformly positive evaluations. Similarly, \citet{djafarova2023exploring} identify review length, writing style, and the inclusion of mixed or two-sided sentiment as key factors influencing credibility perceptions, particularly in travel and tourism contexts. \citet{jakesch2023human} demonstrate that people often associate the use of first-person pronouns, contractions, or references to family with human-written content.

The valence of reviews (positive vs. negative) can significantly shape human perception \citep{metzger2015psychological}. This is supported by framing theory, which suggests that negative reviews tend to carry more weight and are often judged as more credible than positive ones \citep{levin1988consumers, doh2009consumers, mudambi2010research, o2011young, kusumasondjaja2012credibility, maslowska2017too}. Some studies suggest that a disproportionate number of positive online reviews may cause consumers to discount positive reviews as not reliable \citep{chevalier2006effect} and therefore may negatively affect sales. This negativity bias reinforces the idea that consumers may be more sceptical of highly favourable reviews. A study published by the UK DBT in 2023 \citep{DBT2023FakeReviews} found that at least one in ten product reviews on third-party e-commerce platforms are likely to be fake — most of them positive and intended to influence consumer purchase decisions \citep{news2025}. In this context, a fake review was defined as one that does not reflect a genuine experience of the product or service and is designed to mislead consumers. Such reviews can be either incentivised and human-written or generated by AI systems, differing from the focus on  AI-generated reviews alone taken herein.

\subsubsection{Consumer judgment is ‘intuitive yet flawed’}
In \citet{banerjee2017don}, linguistic cues that can distinguish real from fake reviews were summarised into a set of guidelines to help humans better identify deceptive content. While intended as an intervention to enhance human detection accuracy, such findings can paradoxically be repurposed to improve the generation of fake reviews by informing AI systems of which linguistic patterns to avoid to increase perceived authenticity. In other words, these heuristics, while cognitively efficient, make human judgment vulnerable to manipulation, particularly when AI-generated content is designed to mimic these expectations. 

\citet{jakesch2023human} describe these cues as intuitive yet flawed heuristics, demonstrating that they make human judgment of AI-generated language predictable and manipulable, allowing AI systems to produce text perceived as ‘more human than human’. In this sense, human judgment can be systematically biased in favour of well-targeted misinformation, especially when it ‘feels’ intuitively correct - bypassing deeper cognitive evaluation. However, to date there is limited empirical evidence directly confirming whether such heuristics can be intentionally misused to fool human judgment in consumer contexts. This highlights the need for systematic investigation into how both humans and LLMs respond to AI-generated content specifically designed to exploit these cognitive shortcuts.

\subsection{AI judgment of review authenticity}

Prior research on AI-based fake content detection has explored a range of methods, mostly relying on linguistic and textual features such as content length, sentiment, topic distribution, or grammar patterns \citep{crothers2023machine, jabeur2023artificial, wu2025survey}. More recently, with the rise of LLMs, newer detection efforts have been used in these generative models themselves to evaluate whether the content is likely to be real or fake. GenAI is no longer confined to content generation, it is also being used in automatic detection \citep{crothers2023machine}.

That said, LLMs like ChatGPT do not process, and hence `understand' content in a human-like way. Instead they generate or judge text based on a statistical prediction, selecting the most probable next word in a sequence - leveraging structural concepts obtained during training, but with no access to deeper semantics or real-world grounding \citep{lindebaum2024chatgpt}. As a result, their outputs may appear fluent and convincing but can still contain inaccuracies or fabricated details - a phenomenon referred to as `hallucinations’ \citep{alkaissi2023artificial}. This limitation poses challenges for AI-led fake review detection. In particular, LLMs without task-specific fine-tuning often perform inconsistently and unreliably across different types of content \citep{liu2023argugpt}. For example, \citet{salminen2022creating} found that although humans struggled to identify AI-generated fake reviews, a fine-tuned model performed well, highlighting a growing gap between generation and detection capabilities. In contrast, other studies suggest that detection models may fail when the LLMs used to generate the fake content are unknown \citep{kovacs2024turing, wu2025survey}.

These findings underscore a broader concern in the context of customer reviews: while fake reviews can be generated quickly and easily using off-the-shelf LLMs, fake content detection remains an open and urgent problem — especially as LLMs become more powerful and widely accessible \citep{crothers2023machine}. In consumer-facing environments, this imbalance represents a growing risk: the production of persuasive but deceptive content is becoming increasingly scalable, while the capacity to detect it remains limited. This calls for a broader evaluation of AI's role not only as a content generator but also as an emerging judging agent. We contend that it is essential to understand how AI-based authenticity evaluation differs from human judgment. 


\subsection{Human vs. LLM judgment}

\citet{lindebaum2024chatgpt} discussed the fundamental differences between human and machine-based judgment, particularly in tasks involving the evaluation of authenticity and truthfulness. Human judgment is described as involving contextual, reflexive, and deliberative production of meaning, especially in low-probability, ambiguous situations. It is an organic process tied to learning, experience, and moral interpretation within complex socio-technological environments \citep{moser2022morality}. In contrast, LLMs such as ChatGPT do not possess the capacity for reflexive judgment. They rely on statistical prediction to generate the most probable next word in a sequence, without semantic awareness, intuition, or the ability to reason about truth or false claims \citep{hannigan2024beware, lindebaum2024chatgpt}. 

In evaluating content authenticity, sequence length has emerged as an important signal for LLMs, with longer text generally improving detection performance \citep{crothers2023machine}. If this holds true in consumer review contexts, it may suggest that surface cues alone can mislead LLMs into perceiving fake reviews as real. In contrast, human evaluators - drawing on contextual reasoning and critical reflection - often perform better with longer reviews, where more cues are available to support deeper verification. These divergent behaviours underscore the importance of empirically investigating how humans and LLMs differ in processing and evaluating authenticity. This is particularly the case in domains such as online reviews, where trust is central and misjudgment carries critical consequences. However, this remains a relatively underexplored area in existing research.

To empirically examine these challenges, the following sections present three studies comparing human and LLM performance in detecting AI-generated fake reviews.

\section{Study 1: Can Humans Distinguish Between Real and Fake Reviews?}\label{sec:sec3}
In this study, we examined whether humans could distinguish between real (human-written) and fake (LLM-generated) reviews. The analysis utilsed the Amazon Review 2023 dataset\footnote{The Amazon Review 2023 dataset is publicly available at \url{https://amazon-reviews-2023.github.io/}} and
focused on human classification performance, assessing not only overall accuracy but also associated confidence levels, response patterns over time, and relationships with demographic factors when judging reviews.
\subsection{Characteristics of Real Online Reviews from Amazon}\label{sec:real-reviews-characteristics}
The Amazon Review 2023 dataset includes multiple categories, each representing different product types with varying levels of customer engagement \citep{hou2024bridging}. To identify a suitable category for sampling, we ranked them based on key metrics related to customer engagement, product variety, review text volume, and review frequency. ‘Home\_and\_Kitchen’ emerged as the top-ranked category and was thus selected for further analysis.

From this category, we first extracted two sets of authentic human-written reviews to better understand the characteristics of natural review writing. A random sample of 1,000 reviews was used to capture the general patterns, and a balanced sample of 5,000 reviews based on star ratings was used to capture differences across ratings. Insights from this analysis informed the design of prompts for synthetic review generation in the later stage. 


The analysis focused on several key characteristics of customer reviews that have been widely examined in prior research \citep{banerjee2017don,nica2022ill}, particularly those related to diversity and distribution in style and content (see Table~\ref{tab:characteristics}). The main findings are summarised below.

\begin{table}[htbp]
\centering
\caption{Characteristics of reviews and analysis sources.}
\begin{tabular}{lll}
\hline
Characteristics                      & Features                                                  & Source                                    \\\hline
Textual      & Review length  & NLTK 3.9.1 Python library                 \\
                           & Sentence length variance within reviews                   & NLTK 3.9.1 Python library                 \\
                           & Use of uppercase words                                & NLTK 3.9.1 Python library                 \\
                           & Punctuation style                   & Regular expression module in Python 3.11  \\\hline
Linguistic & Use of first-person words       & Regular expression module in Python 3.11  \\
                           & Use of second-person words       & Regular expression module in Python 3.11  \\
                           & Use of past tense verbs                                   & SpaCy 3.7.5 Python library                \\
                           & Use of idiomatic expressions and colloquialisms           & DeepSeek (date 2025-02-05)                \\
                           & Typo, misspelling and grammar mistakes& Language-tool-python 2.8.2 Python library\\\hline

Sentiment & Positive, neutral, negative sentiment       &  VADER sentiment library in NLTK 3.9.1\\\hline

\end{tabular}
\label{tab:characteristics}
\end{table}
\FloatBarrier

\textbf{Review length and structure}: Most reviews were short and to the point, with an average of 30 words and half containing fewer than two sentences. Short reviews (e.g., fewer than 4 words) made up 10\%. Some reviews were much longer, with 10\% exceeding 60 words and the longest reaching 20 sentences. While many reviews had a consistent writing style, some alternate between very short and very long sentences, making them more expressive. 

\textbf{Use of uppercase words}: Most reviews used standard capitalisation. However, 10\% of reviews used uppercase words for emphasis (e.g., ‘LOVE’ or ‘DO NOT’), reflecting strong sentiment. 

\textbf{Punctuation style}: Periods and commas were commonly used, suggesting a neutral tone and well-structured sentence structure. 
About 1\% of sentences contained unusual punctuations such as ellipses ‘...’ and exclamation marks ‘!!!’, reflecting authentic emotional tone. 

\textbf{Pronouns}: First-person pronouns were used moderately, while second-person pronouns were infrequent and typically appeared in instructional or advisory reviews. 

\textbf{Past-tense verbs}: Past tense verbs appeared occasionally (2 per review on average), suggesting some personal experiences or references to past events. Only 12.5\% reviews used more than 5 past-tense verbs, indicating that detailed, story-like narratives are relatively rare. 

\textbf{Idiomatic expressions and colloquialisms}:
Reviews used a mix of informal, expressive language that leads to a conversational tone. Common idioms included ‘with flying colours’ (indicating success), ‘worth every penny’ (emphasising value), and ‘break the bank’ (indicating to be very expensive), reflecting the use of figurative language to convey enthusiasm. Similarly, colloquialisms like ‘lol’ (casual emphasis) and ‘kinda’ (informal hedge) indicated a relaxed and informal writing style.

\textbf{Mistakes}:
The presence of typographical errors, misspellings, and grammar mistakes is important in making reviews feel authentic and human-like, as imperfections are a natural aspect of human writing \citep{bluvstein2024imperfectly}. Unlike machines, which generate perfectly structured text, real users often introduce spelling errors, grammatical inconsistencies, and informal phrasing. 
The sampled reviews showed common issues such as incorrect verb forms, redundant or missing spaces and punctuation, improper capitalisation, missing verbs, and incorrect pronoun usage. Among the common issues, misspellings were the most frequent (in over 50\% of reviews), followed by typographical errors (15\%), and grammar mistakes, whitespace errors, and style issues (collectively around 10\%), while duplications were rare (7 out of 1,000 reviews). 
Higher-rated reviews tended to be more polished, possibly from more engaged or satisfied customers. In contrast, 2-star and 3-star reviews had the highest error rate per sentence, suggesting they are written more casually or quickly. 


\textbf{Sentiment characteristics}: Sentiment generally aligned with star ratings, with higher ratings reflecting more positive sentiment. However, many reviews with mid-range ratings showed mixed sentiment, combining both praise and criticism. 3-star reviews were the most mixed, often including both strengths and weaknesses. 

\subsection{Generating Fake Reviews}\label{sec:fake-review-generation}
To generate human-like reviews using LLMs, prompts should reflect the natural writing patterns of authentic reviews. Based on the previous analysis, we developed heuristics to guide prompt design:
\begin{enumerate}
    \item Review length and structure: Reviews vary widely in length, but most are short. Prompts should permit both brief and detailed responses, emphasising that concise feedback is more acceptable while also encouraging detailed narratives to enhance authenticity.
    \item Capitalisation, punctuation, and expressiveness: Occasional use of uppercase words and informal punctuation for emphasis should be permitted, but not their overuse. Prompts should allow their moderate usage, while avoiding excessive or formal punctuation.
    \item Pronoun usage: Prompts should encourage a product-focused perspective while allowing personal reflections by mentioning first-person experience. Prompts should discourage second-person pronouns unless the context is advisory.
    \item Past-tense usage: Prompts should accommodate both concise feedback and detailed storytelling with past tense to maintain authenticity. Lower-rated reviews reference past experiences more often, while higher-rated reviews focus on current satisfaction.
    \item Idioms, colloquialisms and tone: Prompts should encourage occasional figurative language to enhance conversational and casual tone. Colloquialisms appear in some reviews but are context-dependent, so prompts should indicate their selective use, especially for informal products or experiences.
    \item Common mistakes: Prompts should allow occasional typos, minor inconsistencies, or redundant words, as real users rarely write with perfect accuracy. Misspellings are the most frequent error, followed by typographical and grammar/style issues. Errors should be more common in 2-star and 3-star reviews.
    \item Sentiment: Prompts should align sentiment with ratings. 1-star and 2-star reviews are typically negative, while 5-star reviews are mostly positive. Prompts should encourage a balanced perspective with mixed sentiments, especially in 3-star reviews.
\end{enumerate}


Preliminary testing with simple prompts indicated that certain patterns, such as sentiment and rating alignment, naturally emerge in the model’s output without explicit instructions. 
Therefore, we focused on characteristics not captured by default but critical to perceived authenticity. With these insights, we proposed a set of prompts shown in Table \ref{tab:prompt-update} and used the ChatGPT o1 model to generate reviews grounded in detailed product information. 


\begin{table}[htbp]
\centering
\caption{Proposed prompts and corresponding design insights and heuristics $h$. }
\begin{tabular}{p{10cm} p{4cm} p{0.5cm}}
    \hline
    Prompt & Justification & $h$ \\
    \hline
    Consider the following table: [listing multiple products, each with metadata fields: title, features, description, price, and details.] & Set input context (source data). & \\
    \hline
    Append four new columns: rating (1-5), number of helpful votes (numeric), title (short review title), and text (review content).  
    For each row, generate exactly one review as similar to human-written as possible based on the row’s data. Output the final table with the original columns plus the new ones. & Define expected output and formatting requirements. & \\
    \hline
    Generate ratings following the observed distribution—high ratings should be most frequent, mid-range ratings moderate, and low ratings least common. & Match rating distribution. & \\
    \hline
    Assign helpful votes in line with the real distribution—most reviews should have low vote counts, some moderate, and a few high. & Match vote distribution. & \\
    \hline
    Ensure reviews resemble human writing: vary length, use a casual and relaxed tone, and include minor imperfections like typos and grammatical inconsistencies. & Maintain natural style and authenticity. & \\
    \hline
     Align the sentiment with the rating: make 1-2 star reviews negative without exaggeration, 4-5 star reviews positive, with 5-star being very enthusiastic and 4-star including minor critiques, and 3-star reviews balanced with both praise and criticism. & Guide the emotional tone based on the rating. & 7 \\
    \hline
    Vary review lengths and structures to mimic natural writing: include both brief (1-4 sentence) reviews and longer, detailed ones with varied sentence lengths—even using very short sentences for emphasis. Avoid overly structured or formulaic reviews. & Increase diversity in writing. & 1\\
    \hline
    Introduce natural errors—especially misspellings, typographical mistakes, minor grammatical inconsistencies, and redundant spaces — more often in 2-star and 3-star reviews to enhance authenticity. Aim for an average of five mistakes for every ten reviews. & Mirror the natural imperfections in reviews. &  6 \\    
    \hline
    Vary punctuation style by occasionally using informal markers like ellipses and selective uppercase words for emphasis, while avoiding grammatically rigid and overly formal punctuation like semicolons to maintain a conversational tone. & Mimic punctuation style. &  2 \\
    \hline
    Use second-person pronouns sparingly, mainly for advisory contexts, ensuring the review remains product-centred rather than overly directed at the reader. & Mimic pronoun usage. &  3\\
    \hline
    Use slang, idioms, metaphors, colloquialisms, and figurative language selectively to enhance a natural, casual tone, ensuring they fit the context. & Use appropriate informal expressions. &  5 \\
    \hline
    Incorporate elements of episodic memory or nostalgia to suggest genuine product use, especially when describing issues like breakage or delivery problems. Lower-rated reviews can reference more past experiences, while higher-rated reviews focus on present satisfaction, blending concise feedback with detailed storytelling. & Add depth to the reviews. &  4, 7 \\
    \hline
    The aim should be that a reader cannot distinguish the reviews from standard writing that might be expected on a website like Amazon.com. & Reinforce the main quality benchmark. & \\
    \hline
\end{tabular}
\label{tab:prompt-update}
\end{table}
\FloatBarrier

\subsection{Experimental Method}
\subsubsection{Review dataset}
The study used a total of 50 product reviews \citep[as in][]{data_doi_badge}: 25 authentic reviews randomly sampled from 1,000 Amazon reviews examined in Section \ref{sec:real-reviews-characteristics}, and 25 LLM-generated reviews as described in Section \ref{sec:fake-review-generation}. These two sets of reviews were compared descriptively to ensure consistency. Table \ref{tab:fake-real-reviews} shows they were similar in length and content, with no significant distributional differences confirmed by a Kolmogorov–Smirnov test (p > 0.05).

\begin{table}[H]
\centering
\caption{Comparison of real and LLM-generated reviews in the review dataset. Distributions were assessed using the Kolmogorov–Smirnov test, with corresponding p-values reported in the final column.}
\begin{tabular}{l|llr}
\hline
                             & Real reviews  & Fake reviews   & \multicolumn{1}{l}{}                           \\
\multirow{-2}{*}{Features}   & Mean (STD)    & Mean (STD)     & \multicolumn{1}{l}{\multirow{-2}{*}{p-value}} \\ \hline
Review rating                & 4.4 (1.22)    & 3.92 (1.26)    & {\color[HTML]{1F1F1F} 0.156}                   \\
Helpful vote count           & 0.52 (0.92)   & 1.92 (5.35)    & {\color[HTML]{1F1F1F} 0.915}                   \\
Word count                   & 35.92 (28.89) & 66.44 (100.18) & {\color[HTML]{1F1F1F} 0.156}                   \\
Sentences count          & 2.96 (1.70)   & 4.16 (3.92)    & {\color[HTML]{1F1F1F} 0.710}                    \\
Avg word length/sentence & 44.39 (22.48) & 42.78 (23.38)  & {\color[HTML]{1F1F1F} 0.475}                   \\
Avg words/sentence       & 11.52 (6.03)  & 11.45 (5.87)   & {\color[HTML]{1F1F1F} 0.475}                   \\
Sentence length variation     & 36.08 (78.08) & 14.55 (13.18)  & {\color[HTML]{1F1F1F} 0.915}                   \\
Uppercase word count         & 0.08 (0.28)   & 0.12 (0.44)    & {\color[HTML]{1F1F1F} 1.000}                   \\
First-person pronoun count        & 1.44 (2.18)   & 2.56 (4.87)    & {\color[HTML]{1F1F1F} 0.915}                   \\
Second-person pronoun count       & 0.08 (0.40)   & 0.16 (0.47)    & {\color[HTML]{1F1F1F} 1.000}                   \\
Past-Tense verb count             & 2.00 (2.10)   & 2.28 (4.99)    & {\color[HTML]{1F1F1F} 0.710}                    \\
Writing mistake count               & 1.12 (2.13)   & 0.76 (1.94)    & {\color[HTML]{1F1F1F} 0.915}                   \\
Sentiment polarity          & 0.22 (0.20)   & 0.10 (0.30)    & {\color[HTML]{1F1F1F} 0.285}                   \\ \hline
\end{tabular}
\label{tab:fake-real-reviews}
\end{table}
\FloatBarrier

\subsubsection{Participants}
A total of 300 participants were recruited via Prolific Academic using the platform's representative sampling option, which allows obtaining samples that reflect national population distributions across age, sex, and ethnicity. This study received ethical approval from the University of Nottingham’s Research Ethics Committee. All participants gave informed consent prior to participation and were compensated at a fair hourly rate in line with ethical standards to ensure adequate task engagement. All participants were adults residing in the United Kingdom and fluent in English. No additional inclusion criteria were applied. After excluding 12 participants due to missing values, the final sample consisted of 288 participants (52\% female; mean age of 47).

\subsubsection{Experimental Task}
Participants were informed that the purpose of the study was to investigate whether humans can distinguish between human-written and AI-generated product reviews. Before the task, participants answered two background questions:
(1) highest completed education level, and
(2) familiarity with LLMs rated on a five-point categorical scale (Never used, Heard of it, Occasionally used, Regular user, Use professionally). Although no additional demographic information was collected during the study, demographic data were obtained from Prolific and used in subsequent analyses.

During the test, each participant viewed all 50 reviews, which were presented sequentially in randomised order to control for sequence effects. For each review, participants made a binary judgment (‘Fake’ or ‘Real’) and rated their confidence level on a scale. The full instructions read:

\begin{boxtext}%
{Imagine you are working in Amazon's Quality Assurance Department. Your role is to assess and flag suspicious fake reviews.

You will judge one product review at a time.

On larger screens, product information will appear on the left, while the review will be on the right. Your task is to determine whether the review is human-written (Real) or AI-generated (Fake). Use your best judgment and indicate your confidence level for each decision.

If you are using a mobile device, the layout may adjust-product information and might appear at the top, followed by the review. Regardless of the layout, the task remains the same.

You will read a total of 50 short product reviews, which should take approximately 10 minutes to complete.}
\end{boxtext}

\subsubsection{Measures}
Classification performance was evaluated using accuracy, precision, recall, and the F1-score, which are standard metrics in binary classification \citep{canbek2017binary}. These metrics are derived from the confusion matrix, which includes four outcomes: true positives (TP), false positives (FP), true negatives (TN), and false negatives (FN). As summarised in Table~\ref{tab:metrics}, these outcomes provide the basis for calculating the evaluation metrics. By analysing the confusion matrix, researchers can assess how accurately the model distinguishes between real and fake reviews. In this study, real (human-written) reviews are treated as the positive class, and fake (AI-generated) reviews as the negative. However, we report precision, recall, and F1 scores for both classes to provide a balanced assessment of how well each type of review is identified. Metric definitions are summarised in Table~\ref{tab:metrics}.

\begin{table}[]
\centering
\caption{Definitions of binary classification evaluation metrics.}
\label{tab:metrics}
\begin{tabular}{ll}
\toprule
Term & Definition \\
\midrule
True Positive (TP)   & Real review correctly classified as real \\
False Positive (FP)  & Fake review incorrectly classified as real \\
True Negative (TN)   & Fake review correctly classified as fake \\
False Negative (FN)  & Real review incorrectly classified as fake \\
\midrule
Precision (Real class)& $\frac{TP}{TP + FP}$ — Proportion of reviews classified as real that are actually real \\
Recall (Real class) & $\frac{TP}{TP + FN}$ — Proportion of actual real reviews correctly identified \\
F1-score (Real class) & $2 \cdot \frac{\text{Precision} \cdot \text{Recall}}{\text{Precision} + \text{Recall}}$ — Harmonic mean of precision and recall \\
\bottomrule
\end{tabular}

\vspace{0.5em}
\footnotesize{\textit{Note:} Metrics can also be calculated with fake reviews to assess classification performance.}
\end{table}
\FloatBarrier

We also investigated the relationships between participants’ confidence, review duration (i.e., response time per review), and classification accuracy, as well as how these measures evolved over the course of the task. In addition, zero-order Spearman correlations were used to assess associations between performance and individual-level factors such as LLM familiarity and education. Time series patterns were also examined to explore potential fatigue or learning effects across the 50-review sequence. 

\subsection{Results}
\begin{figure}[htbp]
\caption{Comparison of confusion matrices for human participants and LLMs.}
\centering
\begin{subfigure}[b]{0.48\textwidth}
    \centering
    \includegraphics[height=6cm,width=7cm]{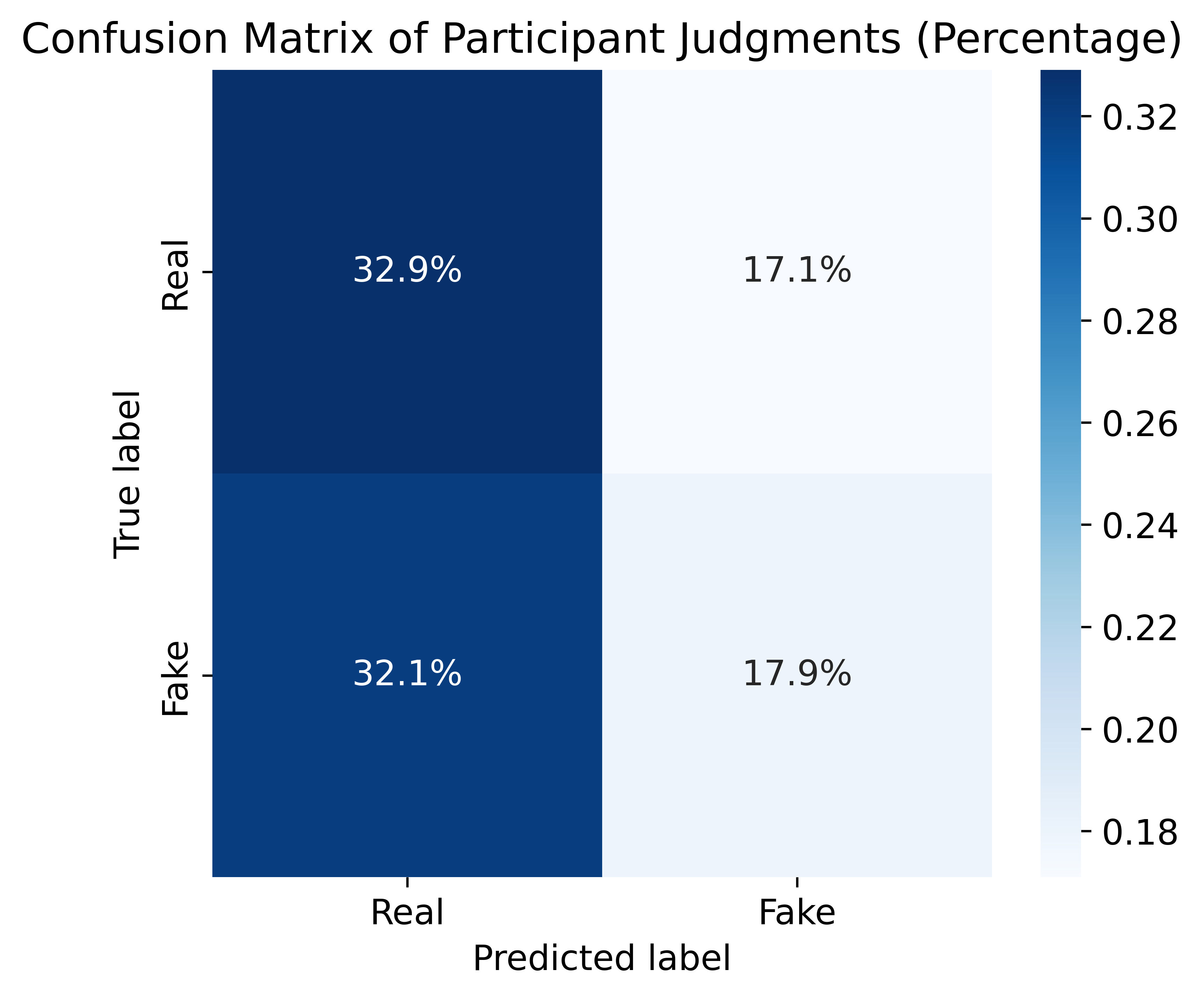}
    \caption{Confusion matrix of participant judgments.}
    \label{fig:cm-human}
\end{subfigure}
\hfill
\begin{subfigure}[b]{0.48\textwidth}
    \centering
    \includegraphics[height=6.25cm,width=7.3cm]{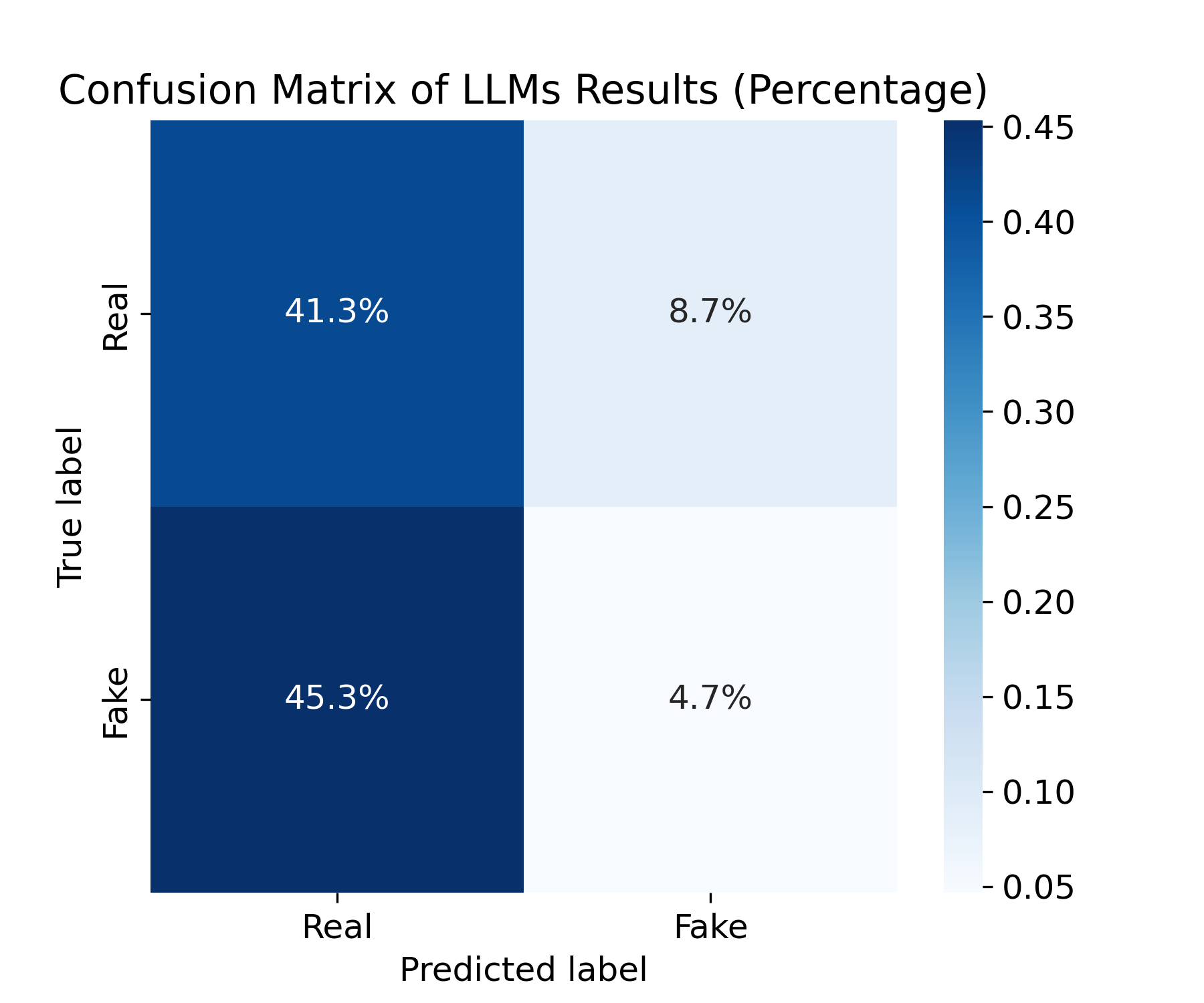}
    \caption{Confusion matrix of LLM results.}
    \label{fig:cm-llm}
\end{subfigure}
\label{fig:cm-comparison}
\end{figure}

\subsubsection{Overall Classification Performance}
As shown in the confusion matrix (Figure~\ref{fig:cm-human}), the overall classification accuracy was 50.82\% (SD = 0.08), which is only marginally above the chance level (50\%). Performance was relatively better when identifying real reviews (precision = 0.506, recall = 0.658, f1-score = 0.572): 65.8\% of real reviews were correctly classified, while 34.2\% of real reviews were misclassified as fake. In contrast, accuracy was substantially lower for fake reviews (precision = 0.512, recall = 0.358, f1-score = 0.421): only 35.8\% were correctly identified as fake, while the majority (64.2\%) were incorrectly labelled as real.

Despite this near-chance performance, participants reported relatively higher confidence in their judgments (Mean = 66.99, SD = 12.03) than their actual accuracy. No significant correlations were found between confidence, time spent per review, and classification accuracy, based on Spearman’s rank correlation tests (p > 0.05 for all pairwise comparisons). 

\subsubsection{Trial-by-Trial Dynamics: Accuracy, Confidence, and Speed}
To explore how participants’ performance and self-perception evolved over time, we examined trial-by-trial changes in accuracy, confidence, and response time across the 50 reviews. Results revealed a slight improvement in accuracy over time, with a modest, statistically significant positive correlation between the trial number and accuracy (r = 0.314, p = 0.026). The observed increase in accuracy appeared to plateau around the 35th review, after which smoothed accuracy remained relatively stable, fluctuating narrowly around 51.5\% and thus, still close to chance.

In contrast, confidence steadily declined as the task progressed, despite the modest gains in performance (r = –0.640, p < 0.001, mean slope of –0.065 per review). Response time also showed a clear downward trend, indicating increased speed over time (r = –0.240, p < 0.001, mean slope of -0.178 per review).

\subsubsection{Demographic and Individual Differences}
\begin{figure}[htbp]
\caption{Relationship between participant age and classification accuracy, with a fitted quadratic trendline to illustrate the negative association.}
\centering
\includegraphics[height=6cm,width=6cm]{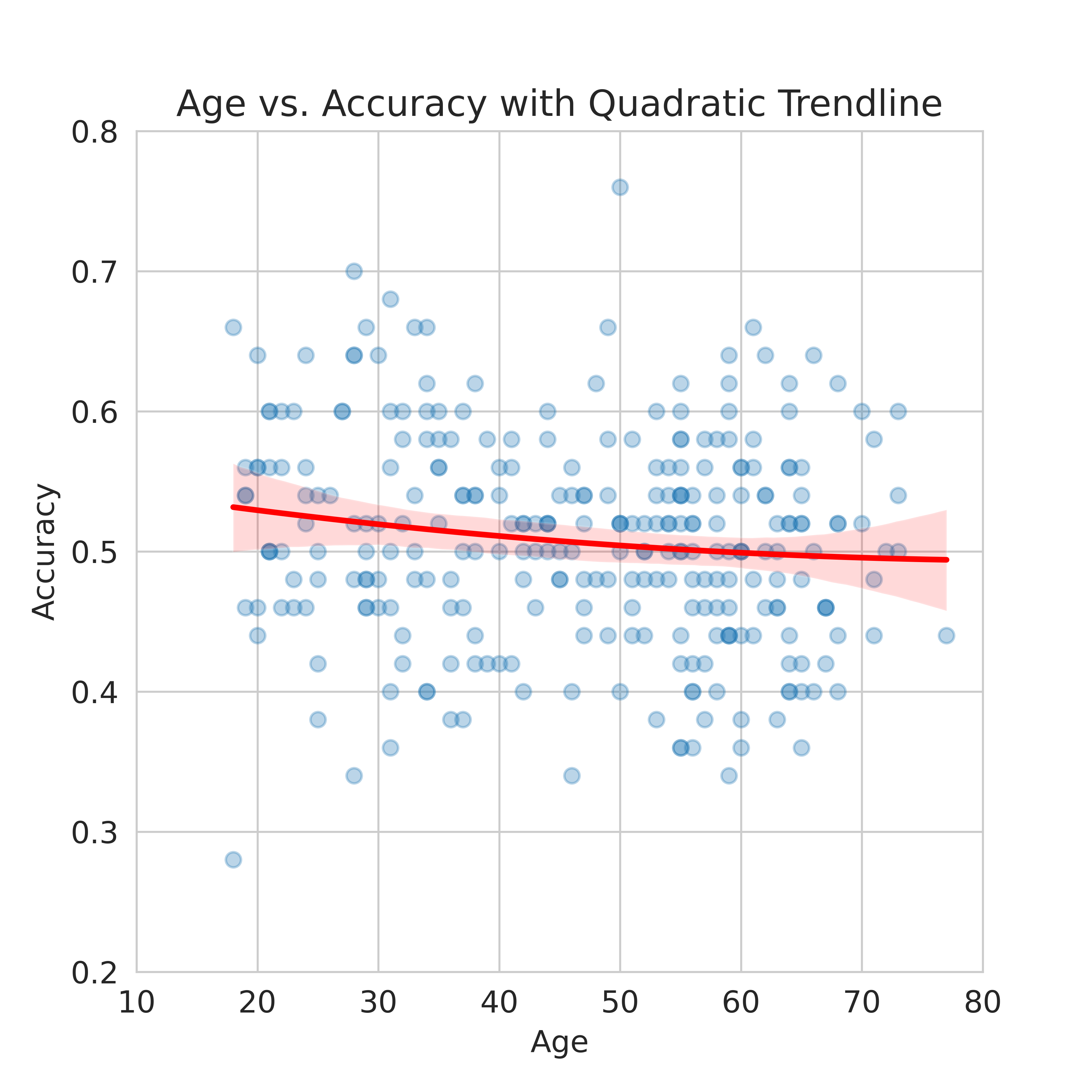}
\label{fig:human-age-accuracy}
\end{figure}

To explore the influence of individual differences on task performance, a correlation analysis was conducted between key outcome measures (accuracy, confidence, and response time) and participant characteristics, including age, gender, number of fluent languages, primary language, education level and self-reported familiarity with LLMs. Age and accuracy had a small but statistically significant negative correlation (r = -0.137, p = 0.019), as shown in Figure \ref{fig:human-age-accuracy}. Age was also positively associated with review completion time (r = 0.284, p < 0.001). However, neither prior familiarity with LLMs, level of confidence, nor any other demographic variables were associated with task performance.

Confidence in task judgments shared no significant relationships with any variables, so neither prior familiarity with LLMs nor educational level, for example, seemed to influence this judgment.

Greater familiarity with LLMs was found to be significantly associated with participants who were younger (r = -0.240, p < 0.001), were fluent in more languages (r = 0.174, p = 0.003), and were more educated (r = 0.234, p < 0.001). Gender was also found to be associated with significant differences in familiarity, whereby males indicated greater levels (M = 2.02) than females (M = 1.55, F(1, 192) = 14.05, p < 0.01).

\subsection{Discussion}
These results reveal a clear asymmetry: participants were moderately better at recognising real reviews but consistently struggled to detect fake ones, with performance near random levels across both classes. This highlights the difficulty of this task for humans - particularly in identifying AI-generated content. Notably, participants demonstrated overconfidence: average confidence ratings were substantially higher than actual accuracy, while confidence showed no significant correlation with performance. If consumers trust their own judgments without sufficient accuracy, they may be more vulnerable to deceptive content.

Further analysis of time-based trends revealed that while participants became slightly more accurate and faster during the task, their confidence steadily declined. Given the absence of feedback during the experiment, participants had limited opportunity to correct their judgments. This may reflect a form of blind learning, where individuals develop rough heuristics but also become increasingly uncertain when faced with ambiguous, hard-to-verify information. Over time, this could reduce confidence in one's ability to assess content authenticity — a risk in real-world environments with extensive AI-generated content.

Individual differences provided limited explanations for their performance. While LLM familiarity was positively related to being younger, more educated, and multilingual, these characteristics did not predict better performance or higher confidence in the task. The only demographic variable associated with accuracy was age, with younger participants tending to perform better than older counterparts.
This advantage, however, could not be fully explained by LLM familiarity alone, suggesting that other factors may play a role. 

\section{Study 2: Can LLMs distinguish between real and fake reviews?} \label{sec:sec4}
While Study 1 focused on human judgment, Study 2 examined the detection capabilities of LLMs themselves. State-of-the-art models, including the one used to generate fake reviews, were assessed to benchmark LLM performance against human intuition.
\subsection{LLM baselines}
In this study, seven leading LLMs were evaluated, including ChatGPT-o1, DeepSeek-R1, Grok-3, Gemini-2.0-Flash-Thinking, ChatGPT-4o, Gemma-3-27B-it, and Qwen2.5-Max. These models were chosen based on three criteria: (1) broad coverage of both open-source and commercial models, (2) high ranking in the Chatbot Arena leaderboard\footnote{\url{https://chat.lmsys.org/}} \citep{chiang2024chatbot} (all within the top 10 as of March 2025), and (3) diversity across model developers. To ensure breadth of representation, no more than one model was included per company. An exception was ChatGPT-o1 as it was used to generate the synthetic reviews in Study 1.
\subsection{Method}
\subsubsection{Task Setup}
The selected models were evaluated on the same 50 reviews used in the human study. The input for each trial included the review text and associated product information, identical to what was shown to human participants, except for product images. All models received the same classification prompt, asking them to determine whether each review was ‘Real’ (human-written) or ‘Fake’ (AI-generated), and to provide a confidence estimate.

All models were queried sequentially (one review at a time) to mirror the structure of the human task. To avoid potential memory effects or conversation history influencing responses, each review was submitted to each model three times using three separate sessions or accounts where possible. This also allowed us to assess the stability and variability of model responses. Due to inconsistencies in how different models control temperature or randomness settings, each model's default generation settings were used to ensure a fair and consistent baseline comparison.

\subsubsection{Measures}
For each model, results were aggregated across the three runs per review. Performance was then compared both between models and against human participants, to determine how effectively current LLMs can distinguish LLM-generated from authentic content. Model performance was assessed using the same metrics as in the human evaluation: accuracy, precision, recall, and F1-score. These metrics were calculated separately for real and fake reviews, allowing for a class-specific performance analysis. 
\subsection{Results: Overall LLM Performance}

\begin{table}[htbp]
\centering
\caption{Overall classification performance of the seven LLMs on the review classification task. Metrics include accuracy, precision, recall, F1-score (aggregated across real and fake classes), mean confidence with standard deviation (SD), and consistency across three repeated trials per model.}
\begin{tabular}{l|llllll}
\hline
Model                     & Accuracy & Precision & Recall & F1-score & Mean confidence (SD) & Consistency out of 3 trials \\ \hline
ChatGPT-o1                & 50.0\%   & 0.752     & 0.507  & 0.348    & 85.93 (2.91)         & 98\%                        \\
DeepSeek-R1              & 38.0\%   & 0.346     & 0.347  & 0.346    & 79.87 (4.47)         & 78\%                        \\
Grok-3                    & 50.0\%     & 0.250     & 0.500  & 0.333    & 84.86 (8.32)         & 100\%                       \\
Gemini-2.0-Flash-Thinking & 48.0\%     & 0.425     & 0.467  & 0.380    & 84.97 (6.30)         & 80\%                        \\
ChatGPT-4o                & 50.0\%     & 0.482     & 0.493  & 0.397    & 90.62 (5.31)         & 78\%                        \\
Gemma-3-27B-it            & 35.6\%   & 0.434     & 0.345  & 0.345    & 87.73 (6.83)         & 80\%                        \\ 
Qwen2.5-Max               & 48.0\%     & 0.430     & 0.474  & 0.377    & 85.36 (7.52)         & 78\%                        \\
\hline
\end{tabular}
\label{tab:llm-performance-overall}
\end{table}

\begin{table}[htbp]
\centering
\caption{Classification performance of the seven LLMs on the review classification task. Precision, recall, and F1-score are reported separately for real and fake review classes.}
\begin{tabular}{l|l|lll}
\hline
Class                 & Model                     & Precision & Recall & F1-score \\ \hline
\multirow{7}{*}{Real} & ChatGPT-o1                & 0.503     & 1      & 0.670    \\
                      & DeepSeek-R1              & 0.358     & 0.387  & 0.372    \\
                      & Grok-3                    & 0.500     & 1      & 0.667    \\
                      & Gemini-2.0-Flash-Thinking & 0.481     & 0.84   & 0.612    \\
                      & ChatGPT-4o                & 0.496     & 0.893  & 0.638    \\
                      & Gemma-3-27B-it            & 0.462     & 0.800  & 0.585    \\
                      & Qwen2.5-Max               & 0.485     & 0.867  & 0.622    \\ \hline
\multirow{7}{*}{Fake} & ChatGPT-o1                & 1         & 0.013  & 0.026    \\
                      & DeepSeek-R1              & 0.333     & 0.307  & 0.319    \\
                      & Grok-3                    & 0         & 0      & 0        \\
                      & Gemini-2.0-Flash-Thinking & 0.368     & 0.093  & 0.149    \\
                      & ChatGPT-4o                & 0.467     & 0.093  & 0.156    \\
                      & Gemma-3-27B-it            & 0.250     & 0.067  & 0.105    \\
                      & Qwen2.5-Max               & 0.375     & 0.080  & 0.132    \\ \hline
\end{tabular}
\label{tab:llm-performance-detail}
\end{table}

In terms of consistency, ChatGPT-o1 and Grok-3 produced nearly identical classifications across repeated trials, with decision consistency rates of 0.98 and 1.00, respectively (in Table \ref{tab:llm-performance-overall}). Other models showed slightly more variability in their responses, with an average consistency rate of 0.79, meaning they produced the same classification in two out of three trials on average.

As shown in Table \ref{tab:llm-performance-detail}, all models performed better at identifying real reviews (Recall-Real scores near 1) but struggled with fake reviews (Recall-Fake scores near 0). DeepSeek-R1 obtained better performance for fake reviews, however, its overall performance in both classes was still poor. ChatGPT-4o, Gemini-2.0-Flash-Thinking, and Qwen2.5-Max showed relatively more balanced performance (highest F1 scores) among the evaluated models (in Table \ref{tab:llm-performance-overall}). The best-performing model, i.e., ChatGPT-4o (accuracy = 50.0\%, F1-score = 0.348), matched human accuracy (50.8\%, F1-score = 0.497) but failed to match their overall effectiveness. 

When comparing model performance as presented in the confusion matrix (in Figure \ref{fig:cm-llm}) with human participants, results showed that humans slightly outperformed all tested LLMs, despite all models showing higher confidence (Mean = 85.62, SD = 3.26). In particular, humans were better at detecting fake reviews as human participants correctly identified 35.8\% of fake reviews, compared to only 9\% for LLMs.


\subsection{Discussion}
The findings highlight a critical limitation of current LLMs: although they can generate highly human-like content by mimicking real reviews, they struggle to recognise such content as AI-generated when prompted to detect it. This suggests that LLMs’ capabilities in generation and detection are not equally developed.

All tested models have a strong bias toward classifying reviews as real. While human participants show a similar tendency, LLMs were more extreme in this regard. As such, it is difficult to consider current LLMs as reliable tools for content authenticity detection — especially given their demonstrated ability to confuse humans with fake reviews.

In this study, human intuition proved more effective than that of the most advanced LLMs. This raises an important question: what underlying cues or heuristics do humans rely on that LLMs fail to capture? This question is addressed in the following section, which explores the key features of reviews that make them particularly difficult or easy to classify, shedding light on the possible sources of divergence between human and model judgments.

\section{Study 3: What makes a fake review hard to distinguish?}\label{sec:sec5}
Having established that both humans and LLMs struggle to reliably detect fake reviews, Study 3 investigated what makes some reviews harder to classify than others by analysing textual, linguistic, and sentiment-related features of the review content. The aim was to uncover the heuristics and cues that humans and LLMs rely on when judging review authenticity and to reveal where their judgment strategies align or diverge.

\subsection{Method}

In this study, Grok-3 was excluded from the analysis as Grok-3 always classified reviews as ‘real’ in Study 2 without showing meaningful variation based on review content.

The features in Table \ref{tab:characteristics} were extracted for the 50 reviews. Initial Spearman correlation analysis showed several features were highly correlated, thus the redundant features were excluded in this analysis. 

To determine what is a difficult review, three indicators of classification difficulty for each review were examined:
\begin{enumerate}
    \item Participant accuracy: the proportion of human participants who correctly classified the review.
    \item LLM accuracy: the proportion of LLMs trials across all models that correctly classified the review.
    \item Participant-LLM similarity: the alignment between human and model classification behaviour.
\end{enumerate}

The Participant-LLM similarity was measured by cosine similarity between their judgment vectors. As each review is either real or fake, judgments can be aggregated and represented by a four-dimensional vector representing classification outcomes, i.e., $[TP, FP, TN, FN]$. This was done separately for human participants and for all trials of the six LLMs. Cosine similarity between the participants and the model judgment vectors was used to quantify how closely the two groups agreed in their classification patterns for each review (higher cosine similarity values indicate higher alignment).

To illustrate how human-model similarity was computed, consider the following example: suppose a particular review is a real (human-written) review. Among the 288 human participants, half correctly classified it as real and the other half incorrectly judged it as fake. The participant classification vector is $[144,0,0,144]$. Across three trials for each of six models (18 judgments total), if 9 classify it as real and 9 as fake. The model classification vector is $[9,0,0,9]$. The cosine similarity between the two 4-dimensional vectors is 1.0, indicating the directional pattern of judgments is the same across humans and models. 
\subsection{Results}


\begin{figure}[htbp]
    \centering
    \caption{Spearman correlations between review features and evaluation metrics}

    \begin{subfigure}[b]{\textwidth}
        \centering
        \includegraphics[height=6cm]{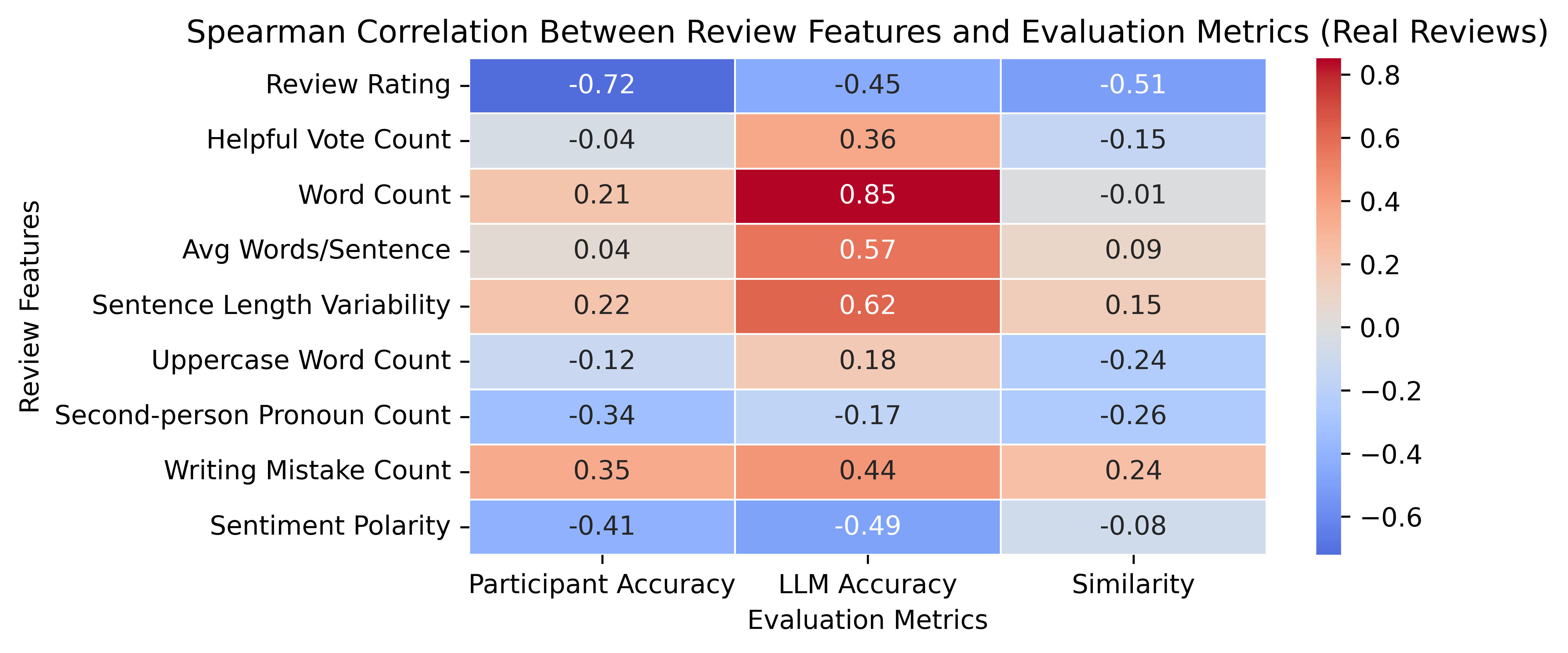}
        \caption{Correlations for real (human-written) reviews.}
        \label{fig:feature-correlation-real}
    \end{subfigure}

    \vspace{0.5cm}

    \begin{subfigure}[b]{\textwidth}
        \centering
        \includegraphics[height=6cm]{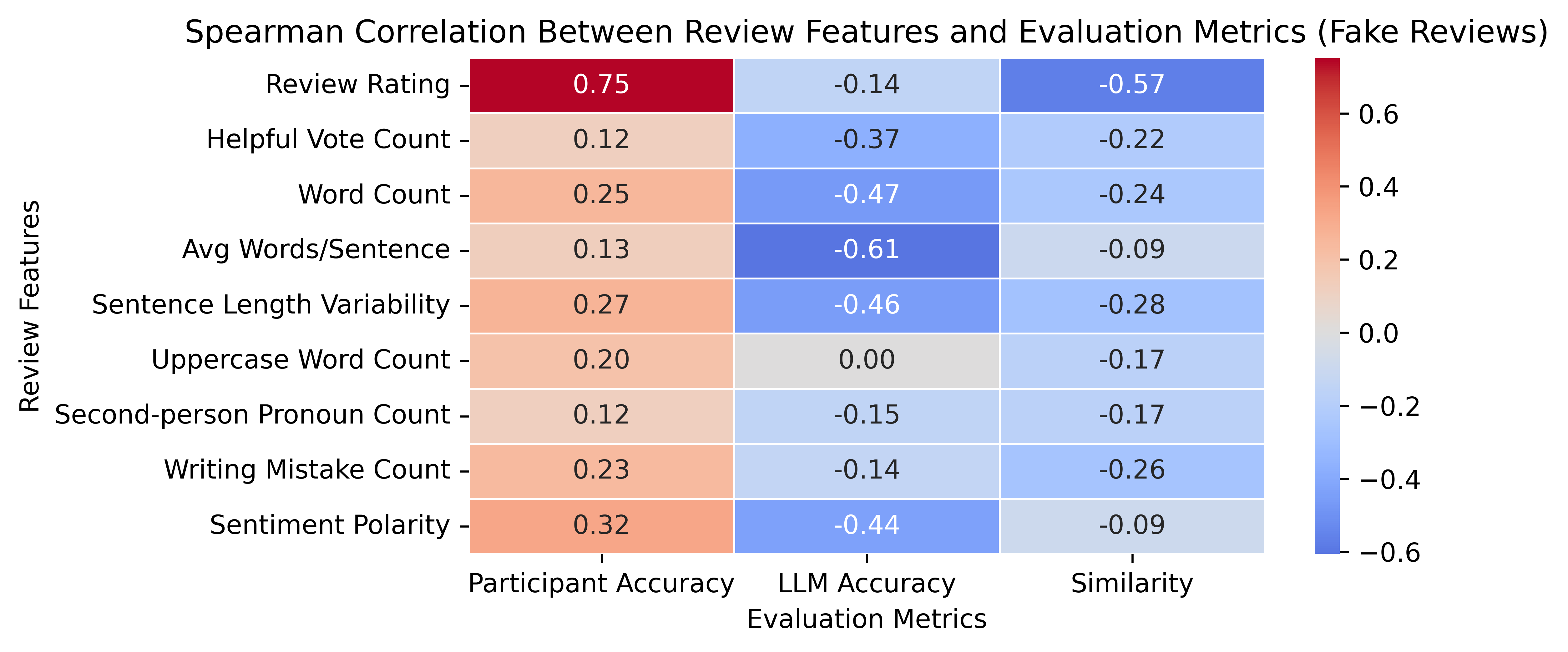}
        \caption{Correlations for fake (LLM-generated) reviews.}
        \label{fig:feature-correlation-fake}
    \end{subfigure}

    \label{fig:feature-correlation-both}
\end{figure}

\subsubsection{Real Reviews: Feature Correlates of Classification Accuracy and Agreement}
When judging real reviews, several text features were associated with the classification difficulty by both participants and LLMs, as shown in Figure \ref{fig:feature-correlation-real}. The review rating was the most influential feature across all three measures. Reviews with higher ratings were significantly harder to classify for both participants (r = -0.721, p < 0.001) and LLMs (r = -0.445, p = 0.026). 

In addition to the decrease in accuracy associated with review rating for both groups, the alignment between human and model judgments was lowest for high-rating reviews (r = –0.506, p = 0.010). This reduced agreement suggests that humans and LLMs made systematically different judgments on these reviews. Specifically, human participants classified more high-rating reviews as fake compared to LLMs.

In terms of review length, LLMs performed better on longer reviews as word count is strongly positively correlated with model accuracy (r = 0.852, p < 0.001). A similar pattern was observed for average words per sentence (r = 0.556, p = 0.003). In addition, reviews with more varied sentence structures were easier for models to classify (r = 0.620, p = 0.001). A similar but non-significant trend was observed for human participants.

For both humans and models, reviews containing more writing mistakes were easier to classify (humans: r = 0.351, p = 0.086; LLMs: r = 0.441, p = 0.027). However, strong positive reviews were associated with lower accuracy for both groups (humans: r = -0.407, p = 0.043; LLMs: r = -0.486, p = 0.014).

\subsubsection{Fake Reviews: Feature Correlates of Classification Accuracy and Agreement}


Unlike in the real review condition, human-model alignment was low across all features, with all correlations negative as shown in Figure \ref{fig:feature-correlation-fake}. Across all features, the review rating showed the strongest associations with participant accuracy. Higher-rated fake reviews were significantly easier for humans to identify (r = 0.751, p < 0.001). At the same time, review ratings were associated with significant misalignment between human participants and LLMs (r = –0.573, p = 0.003).

LLMs show reduced accuracy on longer fake reviews, with negative correlations for word count (r = –0.465, p = 0.019), average words per sentence (r = –0.606, p = 0.001) and sentence length variability (r = –0.456, p = 0.022). In addition, stronger positive sentiment in fake reviews was associated with reduced model accuracy (r = –0.436, p = 0.029).

\subsection{Discussion}
Results indicate several challenges across humans and LLMs in distinguishing real and fake reviews. Review rating, sentiment polarity, the presence or absence of writing mistakes and surface-level richness (e.g., longer length, more detail) are the key diagnostic signals of authenticity for both groups. However, the general human–model agreement remained low, especially for judging fake reviews, suggesting that the two rely on different cues. Overall, these findings point to a slightly different view of how humans and models interpret ‘perfection’ and surface-level richness (e.g., longer length, more detail). 

Humans may exhibit a ‘scepticism bias’ when evaluating the authenticity of online reviews. That is, reviews that are highly rated, strongly positive, and flawlessly written are often treated as fake — possibly triggering the intuition that the review is too good to be true. Therefore, highly polished real reviews may be misclassified as fake, while slightly flawed fake reviews with a more neutral tone and ordinary ratings may be judged as real. This bias may stem from consumers’ knowledge or sensitivity to overly strategic or promotional language. LLMs exhibited a similar trend but with a stronger reliance on surface-level richness, i.e., treating detail-rich and structurally complex content as more authentic. This can be a potentially useful heuristic when judging real content, but misleading when facing fake inputs that are artificially verbose.


There are many differences between human assessors and LLMs, with the former acquiring language in rich, social, and multimodal contexts, while the latter are trained on vastly larger but purely textual corpora \citep{trott2023large}. As a result, human behaviour is shaped by a combination of social, cognitive, emotional, and contextual factors \citep{ecker2022psychological}. In contrast, LLM behaviour is shaped by patterns learned from the large volumes of text on which they were trained. This fundamental difference can explain why LLMs perform especially poorly in detecting fake reviews.


First, some LLMs exhibit what we term a ‘veracity bias’ - a tendency to falsely believe in the veracity of reviews.
For example, models like Grok-3 consistently default to classifying reviews as real. From a game-theoretic perspective, this behaviour can be seen as a form of base-rate optimisation, wherein an agent under uncertainty maximises expected accuracy by aligning with the most frequent class. Fundamentally, this strategy arises from the underlying distribution of training data, which contains predominantly human-generated content. 

Second, other models, while not exhibiting this default tendency, still failed to detect deception reliably — suggesting that they have not developed effective heuristics for identifying AI-generated content. These models may produce more balanced outputs, yet rely on superficial textual cues and lack contextual reasoning. This is reflected in the divergent cue usage between LLMs and human participants.

\section{General Discussion}\label{sec:sec6}


This paper offers empirical evidence that LLMs struggle as much as humans in detecting fake AI-generated reviews. Study 1 highlighted that humans operate at a chance level of detecting fake reviews, with confidence not ensuring accuracy \citep[as in][]{double2024confidence,litwin2025reporting}. The only subtle determinant of higher accuracy was younger age, suggesting that familiarity with AI could play a role. Humans were also prone to be overconfident in their ability to reveal fake content, raising significant concerns regarding the ‘dark side’ of AI in consumer decision-making. As AI-generated text advances and further blurs the boundary between authentic and synthetic reviews, it erodes trust and exploits consumers' vulnerability. Interestingly, LLMs in Study 2 did not perform better than humans. A chance-level accuracy was achieved but for a different reason - the majority of LLMs were selecting ‘real’ for any review. 
LLMs 
almost always classified content as authentic, thereby avoiding false accusations while failing to enhance detection capabilities. Study 3 aimed to unpack which of the review text characteristics were used by humans and LLMs to detect fake reviews. While humans used similar heuristics to those used by our research team in designing the studies, the intuition of LLMs remained a black box. The striking difference between humans and LLMs was found in the perception of what chiefly constituted an authentic review: indications of slight flaws for humans, and verbosity for LLMs. 


Fake reviews posed significant challenges to the credibility and reliability of online marketplaces even before the emergence of ChatGPT, distorting consumer perceptions and undermining genuine feedback \citep{sahut2024antecedents}. Now, with the widescale incursion of AI-generated content, it is no longer possible to establish the ground truth of what is truly authentic. Consumers themselves could be using Gen-AI to create their authentic reviews and articulate their experiences more effectively. However, the large-scale injection of AI-generated reviews poses serious risks, whether intended to promote a brand positively or to harm a competitor’s reputation. Open web data have become polluted sources that seemingly align with the concept of increasing platform ‘enshittification’ \citep{doctorow2023tiktok}. Insights drawn from online reviews that include undetectable AI-generated content could be compromised, ultimately leading to a loss of confidence in users feedback as a decision-making tool. Constant usage of corrupt datasets, by extension, might affect consumer behaviour research in general, exacerbating the cycle of data degradation.


If AI-generated content cannot be reliably detected by either humans or LLMs, the focus should be on developing tools to ensure verification and transparency at least in the future open web data landscape. Digital identity verification could be one of the possible interventions \citep{shukla2024fighting}. Another solution could be watermarks embedded into either or both AI- and human-authored content, ensuring that authenticity can be verified at the source rather than through post-hoc detection efforts. Unfortunately, current advances in watermarking do not work on very short pieces of text, such as reviews \citep{dathathri2024scalable}. However, metadata tagging for the AI-generated review could provide transparency using the platforms to disclose when the review has been synthesised. It goes without saying that these technical solutions must be complemented by ethical and regulatory frameworks ensuring responsible deployment, while maintaining trust in digital communication. 


\section{Conclusion and Future Directions}\label{sec:sec7}
\subsection{Conclusions}
This research investigates the ability of both human participants and state-of-the-art LLMs to distinguish authentic and LLM-generated product reviews through three studies. The findings show both humans and LLMs perform only marginally above chance, with LLMs underperforming relative to human participants. This highlights a critical imbalance between the generative and detective capabilities of current AI systems and the practical challenge of identifying LLM-generated content in consumer-facing digital environments.

Furthermore, this study examines the textual features associated with detection performance and reveals distinct heuristics employed by humans and LLMs. Humans tend to scepticise overly positive reviews while LLMs tend to judge based on surface-level linguistic features. These findings suggest that humans' cognitive intuition might be misused to produce more persuasive and less detectable fake content.

The findings of this study contribute to a deeper understanding of the intersection between consumer behaviour and GenAI. They highlight an urgent need for the development of more robust evaluative frameworks that account for the practical challenges faced by consumers. Such frameworks should aim to refine the online ecosystem in which consumers can make informed decisions based on trustworthy information, while still benefiting from the productive and efficiency-enhancing applications of GenAI. Addressing these challenges will require sustained collaboration between consumer behaviour scholars and AI practitioners to ensure the integrity and transparency of digital information environments.

\subsection{Limitations and future directions}
This study adopted a heuristic-guided prompting approach for fake review generation using a single LLM, which ensured alignment with patterns observed in real human-written reviews. Future research may benefit from using diverse generation strategies, including multiple LLMs with different parameter settings, to assess whether variation in generation strategy affects their creation.

Regarding fake review detection, this study employed standardised prompting to ensure comparability across models and human participants. While the main focus was not on optimising model performance, it is important to acknowledge that LLM outputs can be sensitive to prompt design and parameters (e.g., temperature). Subsequent studies could systematically explore how such factors influence detection performance, potentially offering deeper insights into model-specific reasoning processes.

Beyond these methodology considerations, findings in this study highlight several future research directions worth exploring. First, it is important to understand how the use of GenAI affects consumer behaviour, particularly in terms of trust and purchase decisions. Previous studies show AI-generated reviews might undermine consumer trust \citep{xylogiannopoulos2024chatgpt} and their purchase (booking) intentions \citep{jia2025unpacking}. Future studies could further examine whether these effects vary across product categories, such as between utilitarian and experiential goods, or between high- and low-priced items. Addressing these questions can help identify when and where AI-generated content has the strongest influence on consumers, and in turn, guide the development of more targeted interventions or regulatory responses.

Another direction worth exploring in future studies is whether the detection capability of humans varies across languages and cultural contexts. Language not only facilitates communication but also reflects cultural context, influencing how consumers interpret online content and form trust judgments \citep{alcantara2017language}. Given that linguistic patterns and consumer heuristics differ internationally, exploring cross-cultural differences may reveal new cues or biases affecting both human and algorithmic detection. 

Additionally, future work should explore the development of intelligent mechanisms that enhance human detection capabilities. These may include more intelligent, task-specific detection models, or hybrid frameworks that integrate human judgment with algorithmic assistance. Such ‘human-in-the-loop’ approaches may offer a promising path toward mitigating the limitations identified in both human and LLM performance.



\section*{Acknowledgments}
This research was partly funded by Innovate UK through Knowledge Transfer Partnership (KTP) No. 13737.

\section*{Ethics Statement}
The ethics committee at the University of Nottingham has approved this study, with approval number 202420022.

\section*{Data Availability Statement}
The data that support the findings of this study are openly available at https://doi.org/10.5281/zenodo.15185816, reference number 15185816.

\section*{Financial disclosure}

None reported.

\section*{Conflict of interest}

The authors declare no potential conflict of interests.

\bibliographystyle{apalike}

\bibliography{cas-refs}





\end{document}